\documentclass[10pt,conference]{IEEEtran}

\newif\ifarxiv
\arxivtrue

\newif\ifieee
\ieeefalse

\ifieee
\else
\usepackage[colorlinks=true,allcolors=black]{hyperref} %
\fi

\usepackage{cite}

\ifCLASSINFOpdf
   \usepackage[pdftex]{graphicx}
   \graphicspath{{figs/}}
   \DeclareGraphicsExtensions{.pdf,.jpeg,.png}
\else
   \usepackage[dvips]{graphicx}
   \graphicspath{{../figs/}}
   \DeclareGraphicsExtensions{.eps}
\fi

\usepackage[cmex10]{amsmath}

\usepackage{amsthm}

\usepackage{algorithmic}

\usepackage{array}
\ifCLASSOPTIONcompsoc
  \usepackage[caption=false,font=normalsize,labelfont=sf,textfont=sf]{subfig}
\else
  \usepackage[caption=false,font=footnotesize]{subfig}
\fi
\usepackage{url}

\usepackage{xcolor}

\newif\iffinal
\finaltrue
\newcommand{\cmtid}{109}

\iffinal
\else
\usepackage[switch]{lineno}
\fi

\usepackage{multirow}

\usepackage{transparent}
\usepackage{tikz}
\ifarxiv
    \newcommand\copyrighttext{%
      \scriptsize Accepted for presentation at SIBGRAPI 2025. The final published version will be available on IEEE~Xplore.}
    \newcommand\copyrightnotice{%
    \begin{tikzpicture}[remember picture,overlay]
    \node[anchor=south,yshift=30pt,xshift=0pt] at (current page.south) {\fbox{\transparent{0.85}\parbox{\dimexpr0.6\textwidth-\fboxsep-\fboxrule\relax}{\copyrighttext}}};
    \end{tikzpicture}%
    }
\else
\fi

\IEEEoverridecommandlockouts
\IEEEpubid{\makebox[\columnwidth]{979-8-3315-8951-6/25/\$31.00~\copyright2025 IEEE \hfill}
\hspace{\columnsep}\makebox[\columnwidth]{ }}

\begin{document}
\title{Exploring Light-Weight Object Recognition for Real-Time Document Detection}

\iffinal
\author{
\IEEEauthorblockN{
Lucas Wojcik\IEEEauthorrefmark{1},
Luiz Coelho\IEEEauthorrefmark{2},
Roger Granada\IEEEauthorrefmark{2},
David Menotti\IEEEauthorrefmark{1}
}

\IEEEauthorblockA{\IEEEauthorrefmark{1}
Federal University of Paraná, Curitiba, PR, Brazil
\texttt{\footnotesize \{lmlwojcik, menotti\}@inf.ufpr.br}}

\IEEEauthorblockA{\IEEEauthorrefmark{2}unico - idTech, Brazil
\texttt{\footnotesize \{roger.granada, luiz.coelho\}@unico.io}}
}
\fi

\maketitle

\ifarxiv
    \copyrightnotice
\else
\fi

\newcommand{\dataset}{\gls*{dataset}\xspace}

\newcommand{\urlDataset}{%
  \iffinal
    \url{https://github.com/BOVIFOCR/iwpod-doc-corners.git}
  \fi
}

\ifarxiv
  \vspace{-3mm}
\else
\fi

\begin{abstract}

Object Recognition and Document Skew Estimation have come a long way in terms of performance and efficiency. New models follow one of two directions: improving performance using larger models, and improving efficiency using smaller models. However, real-time document detection and rectification is a niche that is largely unexplored by the literature, yet it remains a vital step for automatic information retrieval from visual documents.
In this work, we strive towards an efficient document detection pipeline that is satisfactory in terms of Optical Character Recognition (OCR) retrieval and faster than other available solutions.
We adapt IWPOD-Net, a license plate detection network, and train it for detection on NBID, a synthetic ID card dataset. We experiment with data augmentation and cross-dataset validation with MIDV (another synthetic ID and passport document dataset) to find the optimal scenario for the model.
Other methods from both the Object Recognition and Skew Estimation state-of-the-art are evaluated for comparison with our approach. We use each method to detect and rectify the document, which is then read by an OCR system. The OCR output is then evaluated using a novel OCR quality metric based on the Levenshtein distance.
Since the end goal is to improve automatic information retrieval, we use the overall OCR quality as a performance metric. We observe that with a promising model, document rectification does not have to be perfect to attain state-of-the-art performance scores. We show that our model is smaller and more efficient than current state-of-the-art solutions while retaining a competitive OCR quality metric. All code is available at \urlDataset
\end{abstract}

\section{Introduction}

Automatic information retrieval from visual documents has been a key challenge in modern systems across the industry. Various types of documents require State-of-the-Art (SotA) approaches to function well, such as banking apps that require official documents for authentication or document digitizers that create PDF versions of document photos.

Over the years, the document recognition SotA has advanced and tackled problems such as lack of data~\cite{lack} or skewed images~\cite{skew}.
However, these solutions are not always applicable in real-world environments, either due to lack of annotated data from the new document domain, or lack of computational resources required to train large state-of-the-art models. Also, newer OCR models still rely on pre-processing~\cite{easyocr}, even if the system itself is performing such processing.

Our particular research interest lies in mobile-capable OCR from official documents, where the preprocessing pipeline includes the rectification of the input image, a step that significantly improves OCR quality (as we show in Section~\ref{sec:results}). 
Our results indicate that, even with a state-of-the-art OCR approach, this is still important for the best information retrieval possible. In this sense, we work with a delicate tradeoff. 
On the one hand, large models are built for better performance, but require more computational resources. 
On the other hand, smaller models require less processing power, and as such can be run on mobile devices, but frequently yield worse results with less generalization power.
This is true both for the OCR system and for the pre-processing steps.

In this work, we tackle the problem of fast, reliable document rectification as a pre-processing step for OCR in official documents. This is a vital task for real-world use cases, as companies receive a large amount of documents that must be processed in the most time-efficient way possible.
We investigate the performance of some models, from skew  estimation and correction to object detection on NBID~\cite{nbid}. This dataset consists of photos of Brazilian ID cards with synthetic data. Our approach takes IWPOD-Net~\cite{iwpod}, a neural network primarily developed for the task of automatic license plate recognition, tweaks a data augmentation system and applies it for NBID, training the model from scratch.

We perform a series of experiments regarding the data augmentation models, cross-dataset validation and OCR performance to investigate the impact of each variable in our pipeline. The cross-dataset approach is done with MIDV~\cite{midv}, which is a synthetic dataset with ten different document types across IDs and passports from various countries. We show that IWPOD-Net, with the best tuning, is faster and just as good as other state-of-the-art approaches. As such, we contribute to the field by presenting a robust experimental scenario that makes it possible for a smaller network to achieve state-of-the-art results, improving efficiency in real-world scenarios.

The remainder of this work is organized as follows. Section~\ref{sec:related} presents an overview of the state of the art in object detection, document skew correction and OCR systems. Section~\ref{sec:method} presents our chosen dataset, model and approach for tackling our problem. Section~\ref{sec:experiments} presents the experimental protocols we set to investigate our research questions, while Section~\ref{sec:results} presents the results of said experiments. Finally, Section~\ref{sec:conclusions} presents the conclusions to our present work.

\section{Related Work}
\label{sec:related}

The state of the art for OCR today has seen improvements with the development of Large Language Model (LLM) technologies. Newer models such as GPT-4~\cite{gpt4} and Gemini 2.0~\cite{gemini} feature state-of-the-art OCR capabilities in the advent of the Vision-Language Model era (VLM)~\cite{vlm1,vlm2}. While these multi-modal models outperform specialized OCR systems, their large size brings new challenges in terms of inference cost and large adaptation overhead.

Meanwhile, traditional OCR system such as DocTR~\cite{doctr} split the OCR task into two stages: detection and recognition. Popular detection networks include CRAFT~\cite{craft} and ResNet~\cite{resnet}, while recognition is usually done using the CRNN~\cite{crnn}. These models are usually more light-weight, being faster and easier to adapt to new scenarios and script types.

OCR is typically the middle part of the document recognition pipeline, sitting between a pre-processing stage and the entity recognition model. However, LLMs are also able to perform entity recognition, a power that we leverage in our experiments.

For document photos, where the document is seen within the picture, the pre-processing stage usually involves a document detection network and/or a skew corrector. Object detection is a classic computer vision problem to which a myriad of applications have been proposed for a series of specific goals. The YOLO~\cite{yolo} series of models innovated in the fact that a single pass was enough for the entire network's pipeline to run, meaning it was a more efficient model.
The latest model, YOLO11, comes in various sizes and its tiny version is used in this study as an example of small object detector.\footnote{At the time of writing, YOLOv12 is already out - but no paper has been published.}

Another model used in this study is RTMDet~\cite{rtmdet}, which also aims to improve the accuracy / overhead balance through a more compatible design and better training techniques. 
There's also RetinaNet~\cite{retinanet}, which integrates a feature pyramid network~\cite{fpn} as backbone for a two-stage detector. It improves by developing a novel loss called the Focal Loss, which identifies and minimizes class imbalances at training time.

Document skew estimation and correction is another technique used for rectification to achieve better OCR results. In~\cite{jdeskew} we have a state-of-the-art approach using a projection over the Fourier magnitude spectrum that significantly outperforms previous models in terms of runtime. 
We use this model through the Python package provided by the authors, whence came the name we use for this approach: Jdeskew.

Our present work uses a license plate detector called IWPOD-Net~\cite{iwpod}, used for license plates, which are a similar domain to official documents, as explained in Section~\ref{sec:model}. This model is smaller, faster, and its end-to-end OCR performance becomes competitive compared to the state-of-the-art after our experimental fine-tuning.
In Section~\ref{sec:method}, we show our approach for NBID pre-processing using IWPOD-Net.

\section{Methodology}
\label{sec:method}

In this section, we present the target dataset and the model used.
Our main goal is to obtain a small and fast model for document detection on NBID~\cite{nbid}.
This model must also reach a competitive state-of-the-art result in terms of the final OCR.
This model will be included in an information extraction pipeline as the first pre-processing step, aiming at a better information retrieval fidelity at the end (in this case, OCR).
As such, we evaluate the model's effectiveness through the OCR metrics instead of the document detection quality.

\subsection{Dataset}
\label{sec:dataset}

\begin{figure}
    \centering
    \includegraphics[width=0.9\linewidth]{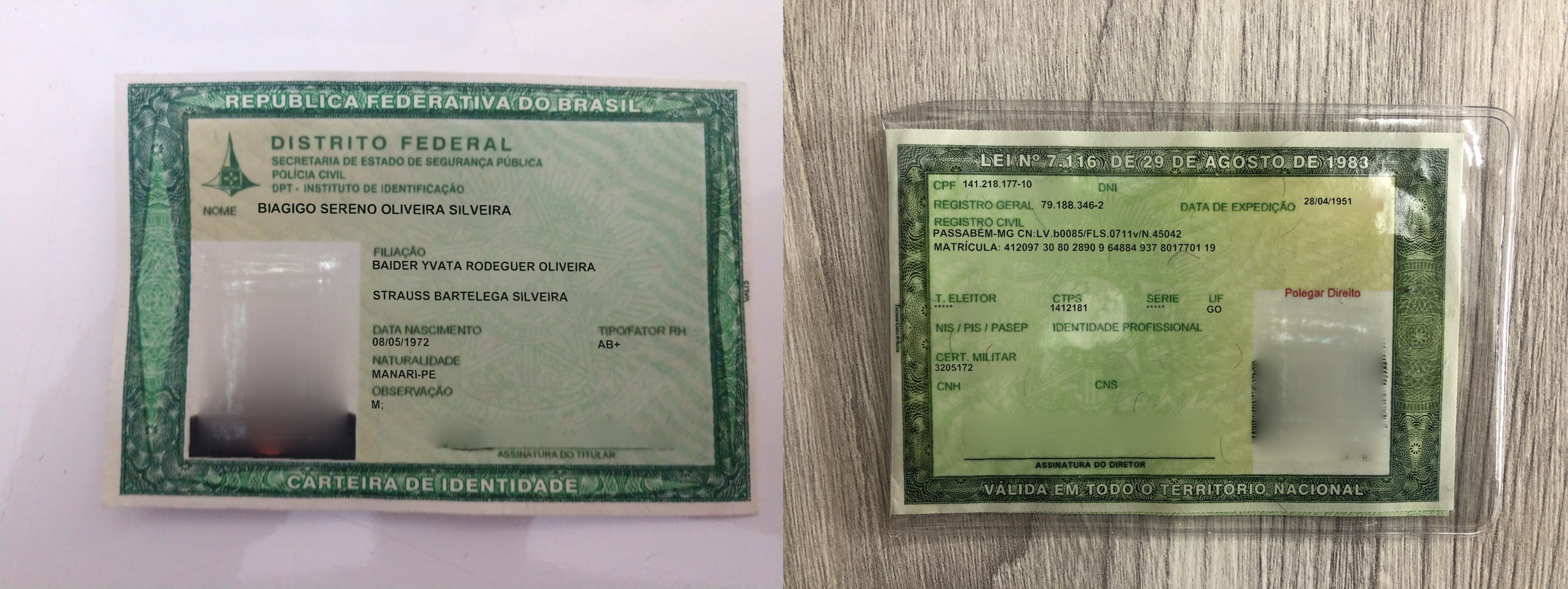}
    \caption{Sample documents from NBID}
    \label{fig:nbid}
\end{figure}

We use NBID~\cite{nbid}, standing for New Brazilian Identity Dataset.
NBID is a synthetic dataset composed of Brazilian ID cards that were inpainted and synthesized for public use. 
NBID authors have made 5 synthetic instances, each with different data, out of each real instance available. In total, there are 1255 synthetic instances crafted from 251 real images. Figure~\ref{fig:nbid} presents two samples from the dataset.

NBID contains a variety of backgrounds and document distortions, and is representative of the images seen in the real world. Although the text inpainting is imperfect, the text is the only element changed in the synthetic instances, with the document object being left untouched. As such, these are good real-world examples as they retain environment variations on lighting, camera quality and document~deformations.

\begin{figure}
    \centering
    \includegraphics[width=0.6\linewidth]{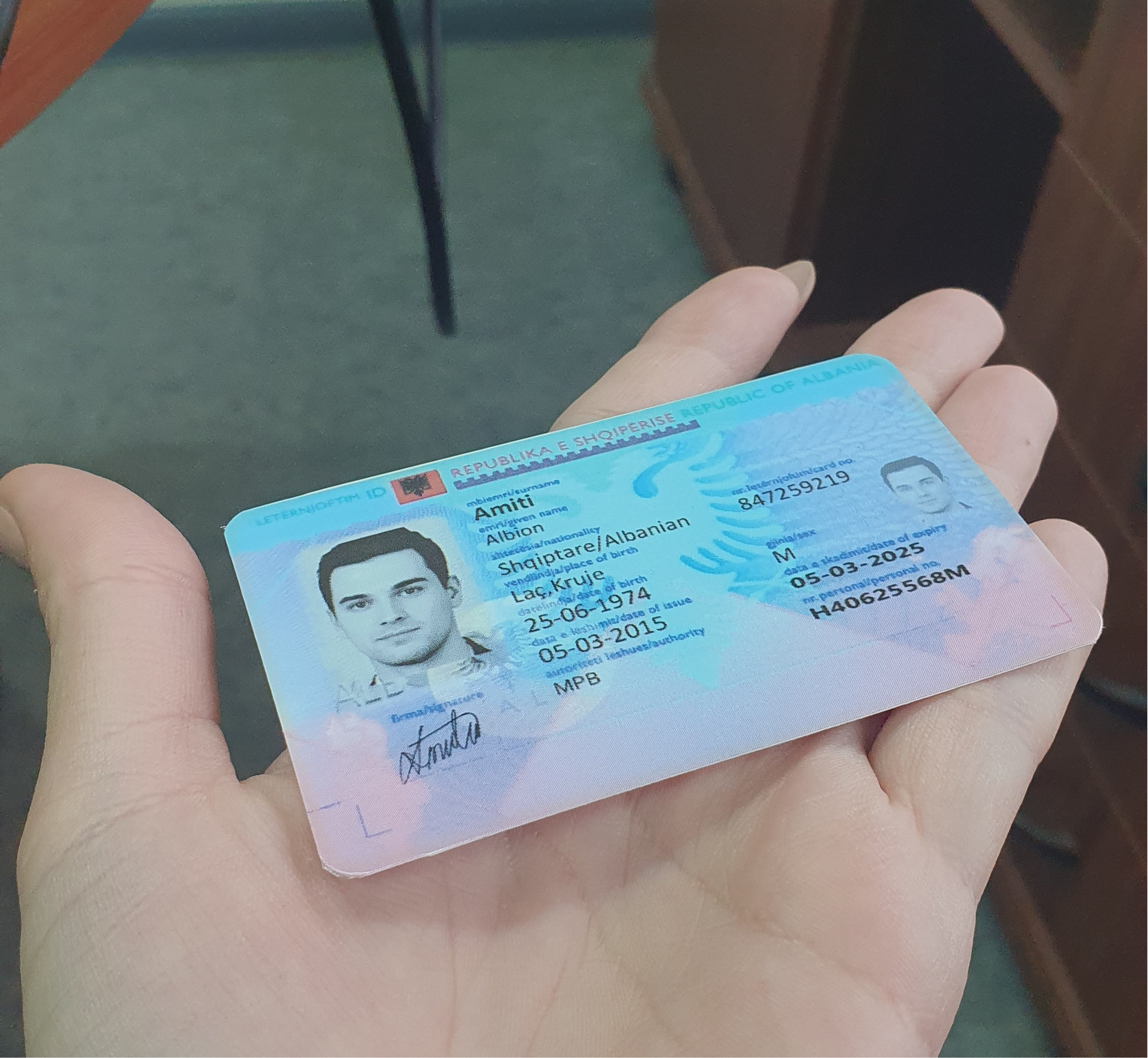}
    \caption{Sample document photo from MIDV}
    \label{fig:midv_sample}
\end{figure}

Here, we work with only one document instance instead of the five synthetic ones per real image.
This is because, since the box detection model does not use the textual cues for learning, it would be redundant to include more than one synthetic instance per document.
We also remove a few instances that featured severe document distortion, seen when a document appears to be too curved, with rectification resulting in a distorted image, such that we are left with 236 instances across front and back documents. Also, we work with MIDV~\cite{midv} for a cross-dataset training and validation experiment. MIDV is another example of the document domain we aim at, featuring a thousand instances across ten document types, five ID cards and five passport types in the photo partition, which is the one we use. A cropped example of one such document instance is presented in Figure~\ref{fig:midv_sample}.

In our experiments, we adopt a 10-fold cross-validation protocol for both NBID and MIDV, splitting the dataset into ten bins, of which one is chosen for validation and another for testing. These bins change iteratively in the usual ten-fold scenario, totaling ten different training rounds. 
\iffinal
These folds were generated by us and are publicly available \footnote{https://github.com/BOVIFOCR/iwpod-doc-corners} for the sake of better reproducibility.
\fi

\subsection{Modeling}
\label{sec:model}

We borrow the IWPOD-Net~\cite{iwpod} model from the license plate recognition literature. IWPOD-Net stands for Improved Warped License Planar Object Detection network, and improves upon the previous WPOD-Net~\cite{wpod} model.
Both are fully-convolutional neural networks that are able to perform single-class, multi-object detection in a single pass.
Also, both networks treat the object detection task by encoding the localization parameters of the object (in this case, the license plate) as a set of six affine transformation parameters that represent the warping of a canonical square into the object polygon within the input image (where it appears distorted).

\begin{figure}
    \centering
    \includegraphics[width=0.9\linewidth]{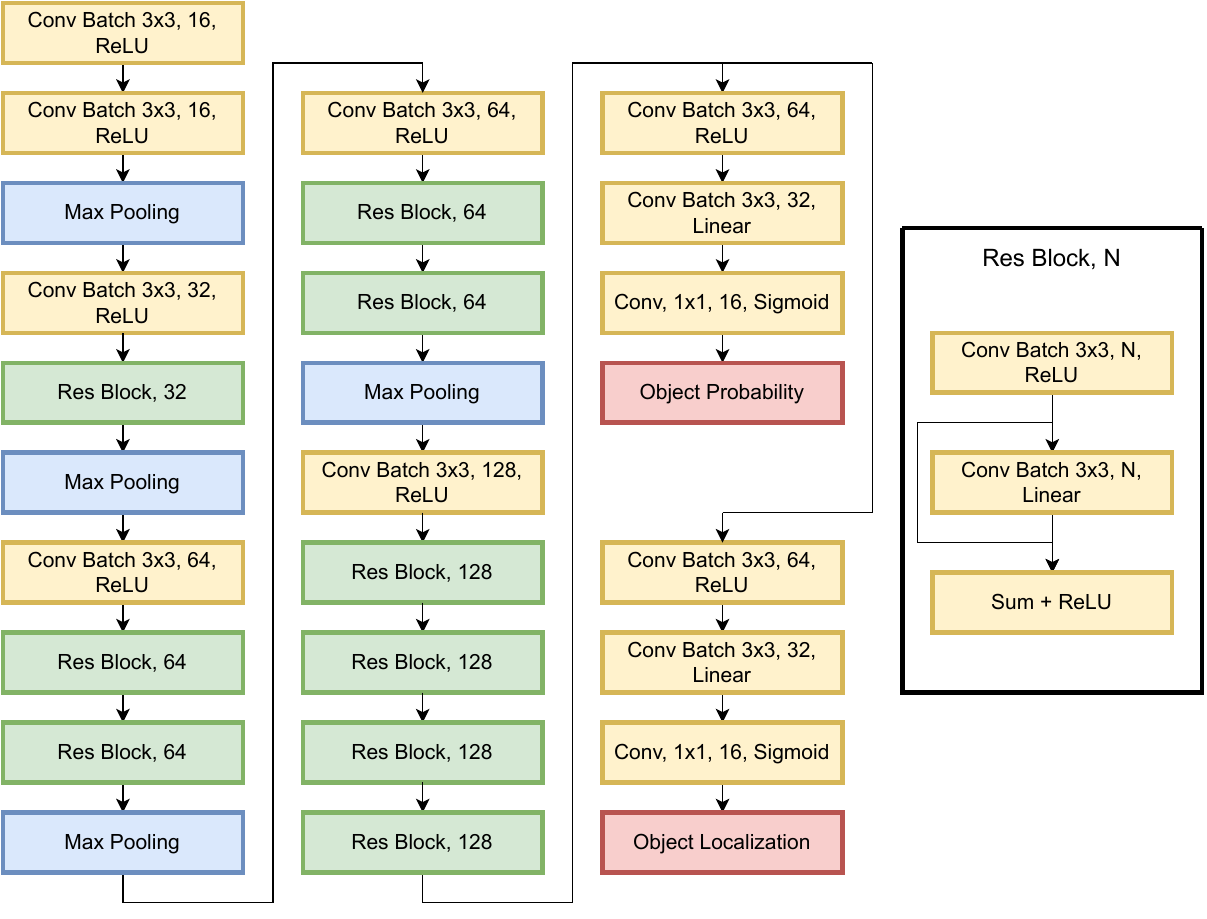}
    \caption{IWPOD-Net Architecture}
    \label{fig:iwpod}
\end{figure}

IWPOD-Net improves upon WPOD-Net by using two sub-networks to treat object probability (detection) and localization separately. These are shallow and independent sequences of convolutional layers for each problem. This is done so that each sub-network stops back-propagating information that might be mutually conflicting when the presentation of input images starts varying a lot in terms of shape and appearance.
This is shown in Figure~\ref{fig:iwpod}, where the full architecture of IWPOD-Net is displayed. This diagram was created by us, based on the illustration present on IWPOD-Net's original paper.

The reason behind choosing IWPOD-Net for our experiments lies in the fact that the treatment of license plates in terms of a canonical, rectified size, translates well to our domain of document detection, since all of the documents contained in NBID~\cite{nbid} feature the same canonical aspect ratio, albeit the rectangle is less stretched. In this sense, IWPOD-Net presents itself as a highly specialized object detection network, tailored for the intricacies of license plates, and can be adapted for documents as these intricacies are somewhat shared in the new domain. While license plates tend to be closer to a parallelogram and documents can be bent and folded, we find that our approach reaches a satisfactory performance level in the detection metric. Also, IWPOD-Net is a small and efficient model, which makes it suitable for running on mobile devices. We analyze its runtime comparative to other methods in Section~\ref{sec:results}.

Furthermore, IWPOD-Net authors also present a robust data augmentation system that is largely domain-agnostic. It is a three-step pipeline consisting of a random crop fully containing the object of interest, a 3D rotation with randomized parameters, and a photometric augmentation that alters the color features of the image. These augmentations are done in an online manner, creating new images at training time.

The random crop is generated by picking new image width and height parameters, rectifying the license plate and selecting random horizontal and vertical offsets to place the license plate onto the new image at a random location. The new object coordinates are also calculated along the process. The random 3D rotation is done by selecting random roll, pitch and yaw parameters within a given threshold, computing the transform matrix and applying it to the image and object coordinates. The original paper uses $\pm 45$\textdegree, $\pm 80$\textdegree~and $\pm 80$\textdegree~for roll, pitch and yaw respectively, but we change these for an ablation study, as explained in Section~\ref{sec:experiments}.

\begin{figure}
    \centering
    \includegraphics[width=\linewidth]{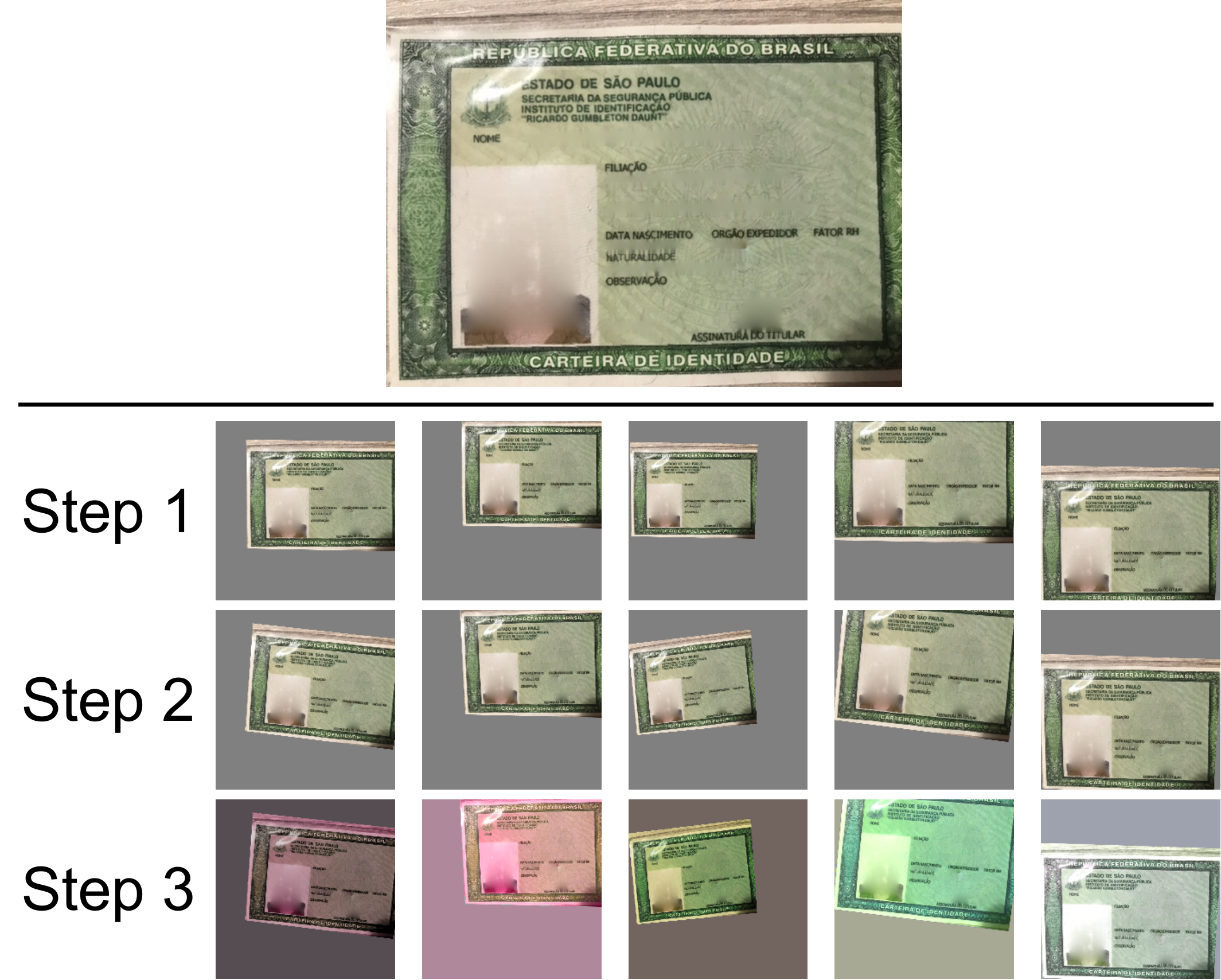}
    \caption{Augmentation Pipeline}
    \label{fig:augmentation}
\end{figure}

The photometric augmentation works by using three different methods: taking the image's negative version, applying Gaussian blur and modifying the HSV colorspace.
From this set of transformations, the image is modified according to a probability: 5\%, 15\% and 100\% respectively for each method. 
We experiment with the photometric augmentation by turning it on or off in our ablation studies. Intuitively, this augmentation method should introduce an undesirable variety in NBID since the documents have a fixed color scheme, but we find that it does improve performance in some scenarios.
Figure~\ref{fig:augmentation} illustrates the augmentation pipeline showing an original image and the resulting augmented images.

\section{Experiments}
\label{sec:experiments}

In order to verify the efficacy of our approach, we perform three sets of experiments. First, we perform an ablation study regarding the data augmentation techniques in order to find the best fit for our dataset. Then, we perform cross-dataset training using MIDV-2020~\cite{midv} to investigate the efficacy of adding more, different training data in regards to performance in NBID. Finally, we use a novel OCR quality metric to compare the final OCR result across different document rectification methods.

It is important to note that NBID contains five synthetic instances per real image used for document augmentation. In this work, we use only one of these documents, meaning our baseline dataset is one fifth of NBID's total size. We work with 236 instances, and exclude a few images from NBID where the document appeared severely distorted. The training, validation and testing dataset splits are done by means of an usual 10-fold separation, each fold serving as validation and testing exactly once across the ten training rounds we perform for the first and second sets of experiments.

The data augmentation method employed in our ablation study is the same as described in Section~\ref{sec:method}, but we change the parameters and enable and disable certain methods. In this case, we change the maximum angle for pitch, yaw and roll for the 3D augmentation, using degree values (\(\sigma\)) of \(15\), \(30\), \(45\), \(55\) and 75. 
For \(\sigma > 45\), the maximum angle for roll was set to \(45\). 
We also include or exclude the photometric augmentation, such that there is two experiment rounds for every value of \(\sigma\). For this experiment, we report the average intersection-over-union (IoU) value between the ground truth polygon and the predicted polygon across all testing instances, averaged again over 10 training rounds.

Second, for our cross-dataset approach, we use MIDV-2020, which is another synthetic document dataset. MIDV is a multimedia dataset of which we use the photo partition of passports and ID cards. In total, it contains a thousand document instances across ten different document types, spanning ten countries. It is a significantly larger sample than the baseline NBID dataset, which features 236 instances in total.

Our cross-dataset training is done on three protocols: intra, cross and multi. Intra-dataset validation consists on testing on the same dataset used for training, and we do this for both NBID and MIDV, separately. Cross-dataset validation consists on testing on the dataset not used for training, that is, the model trained on NBID is tested on MIDV. Finally, multi-dataset validation is done by training on both datasets and testing on both as well. We also use the test-wise mean IoU metric for this experiment, averaged across 10 executions.

Our third and final evaluation protocol is aimed at analyzing the final OCR score. This is our ultimate goal, since a higher OCR fidelity yields better information retrieval. We compare the annotated ground truth text of the document's semantic entities to the corresponding text predicted by the OCR system, for which we picked Gemini-2.0. Gemini is a LLM that has powerful OCR capabilities, yielding scores of over \(0.95\) across most of the methods we used for evaluation. The reason behind this choice is Gemini's state of the art status for general OCR at the time of development of this work, leading to its widespread real-world use. For comparison, and also in order to mitigate model-specific bias, we also use EasyOCR~\cite{easyocr}, a publicly available and open source OCR tool.

Our OCR score is designed around the Levenshtein distance and is normalized between \(0\) and \(1\), where \(1\) means the compared entity texts are identical and \(0\) means they are completely different.
Equation~\ref{eq:dist_score} presents the calculation of the edit distance between the ground truth texts (represented as an array \(GT\)) and the corresponding texts predicted by the OCR system \(PD\). Equation~\ref{eq:ocr_score} presents the final OCR score calculation, where we use the overall edit distance as an inverse measure of quality, i.e., 

\begin{multline}
\label{eq:dist_score}
    Ldist(GT, PD) = \sum\limits_{i=0}^{Len(GT)} min(Lev(GT[i], PD[i]), \\ Len(GT[i]))
\end{multline}

\begin{equation}
\label{eq:ocr_score}
    Score = 1 - \frac{Ldist(GT, PD)}{\sum\limits_{i=0}^{i=Len(GT)}Len(GT[i])}
\end{equation}

Our full OCR evaluation protocol consists of rectifying the document with each one of the compared methods. These are IWPOD-Net, two state-of-the-art rotated object detectors: RTM-Det~\cite{rtmdet} and YOLO11~\cite{yolov11}, one document skew estimator: Jdeskew~\cite{jdeskew}, and two baseline approaches: no rectification and ground truth rectification, where the document is projected using the annotated bounding box.
The object detectors, including IWPOD-Net, function as a two-stage process.
In the first stage, the model detects potential objects within the image, and in the second stage the object with the highest confidence is returned.
The rectified document is inputted into the OCR system, which yields a series of texts that are matched into their corresponding semantic entities in the document, from which the OCR score is then calculated.

All CPU experiments are done on a Ryzen 9 5950X, and all GPU experiments are done on a RTX 3090.
We use the Ultralytics implementation of YOLO11 tiny for oriented bounding box detection (obb) and the mmdet implementation of RTMDet tiny for obb as well. For Jdeskew, we use the python package out of the box. Finally, the implementation of IWPOD-Net that we use is the one provided by the authors. The inference batch size is always 1.

\section{Results}
\label{sec:results}

\begin{table}
    \centering
    \renewcommand{\arraystretch}{1.3}
    \caption{Data augmentation results for 10-fold NBID. We report the mean IoU score across each partition.}
    \begin{tabular}{|c|c||c|c|c|} \hline
    \multirow{2}{*}{\shortstack[c]{3D Transform \\ Max Angle}} & \multirow{2}{*}{\shortstack[c]{Photometric \\ Augmentation}} & \multirow{2}{*}{Train} & \multirow{2}{*}{Validation} & \multirow{2}{*}{Test} \\ 
     &  & & &  \\ \hline
\(\sigma=0\) & No & 97.58 & 87.51 & 89.3 \\ \hline
\(\sigma=15\) & No & 91.04 & 88.79 & 90.34 \\ \hline
\(\sigma=15\) & Yes & 91.9  & 89.94 & 90.79 \\ \hline
\(\sigma=30\) & No & 93.58 & 91.99 & 93.97 \\ \hline
\(\sigma=30\) & Yes & 93.62 & 92.15 & 94.21 \\ \hline
\(\sigma=45\) & No & 93.67 & 91.89 & 94.08 \\ \hline
\(\sigma=45\) & Yes & 92.98 & 92.55 & 94.46 \\ \hline
\(\sigma=55\) & No & 95.27 & \textbf{94.3}  & 95.68 \\ \hline
\(\sigma=55\) & Yes & 93.18 & 92.25 & 94.29 \\ \hline
\(\sigma=75\) & No & 95.1  & 93.05 & 95.05 \\ \hline
\(\sigma=75\) & Yes & 94.87 & \underline{93.84} & 95.28 \\ \hline
    \end{tabular}
    \label{tab:aug_ablation}
\end{table}

Table~\ref{tab:aug_ablation} presents our results for the data augmentation ablation studies. As previously stated, these results are an average of ten training executions performed with a 10-fold split of the dataset.
We report the IoU metric averaged across each dataset split. 

As the results show, the photometric augmentation generally improves performance, except for $\sigma = 55$, where the performance drops across all partitions. 
This scenario also features the best IoU metric, reaching a value of $94$\% on validation. 
Also, even though NBID does not feature wildly distorted documents, we can also see that increasing the distortion level adds such variety that benefits overall performance, even at high distortion levels with $\sigma = 55$ and $75$. 
Finally, we can also see that using any augmentation method is better than not using it when it comes to overall performance.

The first row of Table~\ref{tab:aug_ablation} presents the result for training without any data augmentation, and the results show that the model faced overfitting, with the best result for training but the worst for validation and test.

\begin{table}
    \centering
    \renewcommand{\arraystretch}{1.3}
    \caption{Cross-dataset results for 10-fold NBID and MIDV. We report the mean IoU score across each partition.}
    \begin{tabular}{|c||c|c|c|} \hline
    Protocol & Train & Validation & Test \\ \hline\hline
MIDV Intra & 94.06 & 93.88  & 93.71  \\ \hline
NBID Intra & 95.27 & \underline{94.3} & 95.68  \\ \hline
MIDV Cross & 75.22 & 74.03  & 76.4 \\ \hline
NBID Cross & 63.08 & 61.92  & 60.64  \\ \hline
MIDV Multi & 93.91 & 93.5 & 93.68 \\ \hline
NBID Multi & 94.97 & \textbf{95.41}  & 95.16   \\ \hline
    \end{tabular}
    \label{tab:cross_dataset}
\end{table}

Table~\ref{tab:cross_dataset} presents our results on the NBID-MIDV cross-dataset experiment. For the intra protocol, we train and validate on the same dataset used in test, the cross protocol means testing on the dataset not used for training and validation --- in this case, NBID Cross means training and validating on NBID and testing on MIDV (and vice-versa), and multi means training and testing on both datasets. Again, we report the IoU score averaged across the entire dataset partition over a 10-fold cross-validation protocol. We use the augmentation approach that yielded the best results on validation in our ablation study, meaning \(\sigma = 55\) and no photometric augmentation.

Overall, we can see that joint training slightly improves the NBID performance on validation, but this is not seen for MIDV, where the result is slightly worse. We can also see that NBID performs slightly better in the cross-dataset scenario compared to MIDV. This might be because the data augmentation method was tailored for NBID specifically, and as such it should possess better generalization capabilities. 

The results also show that neither dataset possesses strong generalization potential, as both Cross experiments yielded a poor performance.
We highlight that the same models from the Intra training rounds are used for Cross testing by changing the partitions for the corresponding folds of the other dataset. As such, the Train result for MIDV Cross means that we use the models trained on MIDV Intra and test them on the training partitions of NBID. The same holds for all Cross results.

We also highlight that the data augmentation employed made it possible for the model to recognize documents on both cross-dataset scenarios.
A similar Cross experiment was performed using the models trained without any data augmentation, and for both the NBID and MIDV Cross scenarios we observed a failure in document detection for the test dataset, with less than $10$\% documents being detected, something that was not seen in other experiments.

\begin{table}
    \centering
    \renewcommand{\arraystretch}{1.3}
    \caption{Comparison results across all the collected models on NBID. Time is given in milliseconds, unless stated otherwise. Size is given in number of parameters.}
    \begin{tabular}{|c|c||c|c||c|c|c|} \hline
        \multirow{2}{*}{Model} & \multirow{2}{*}{Size} & \multirow{2}{*}{\shortstack[c]{CPU \\ Time}} & \multirow{2}{*}{\shortstack[c]{GPU \\ Time}} & \multirow{2}{*}{\shortstack[c]{IoU \\ Score}} & \multirow{2}{*}{\shortstack[c]{Gemini \\ Score}} & \multirow{2}{*}{\shortstack[c]{EasyOCR \\ Score}}   \\
        & & & & & & \\ \hline \hline
        None & - & - & - & - & 89.04 & 88.33 \\ \hline
        GT Box & - & - & - & 100 & 97.43 & 89.8 \\  \hline \hline
        IWPOD & 1.8M & 12.56 & 4.96 & 95.53 & 97.73 & 89.21\\ \hline
        YOLO11  & 2.6M & 50.74 & 6.38 & 89.61 & 97.17 & 88.67 \\ \hline
        RTMDet & 4.9M & 1.24s & 18.04 & 93.7 & 97.16 & 89.2 \\ \hline
        Jdeskew  & - & 1s & - & - & 97.73 & 89.32 \\ \hline
    \end{tabular}
    \label{tab:ocr_results}
\end{table}

Finally, Table~\ref{tab:ocr_results} presents our results on the OCR experiment. Here, we take the average OCR score across the entire dataset, with the documents being rectified using each approach listed in Section~\ref{sec:experiments}. We use the Gemini 2.0 model for OCR and entity recognition on a single run. In this case, our prompt asks the model to not only provide the text but also match it to the entity type. We also use EasyOCR~\cite{easyocr}, a public and open-source tool. For EasyOCR, we match each predicted text to the annotated entities according to the highest bounding box IOU. We also present the mean testing IoU score for the object detection networks and the runtime on CPU and GPU. 
As previously stated, the CPU is a Ryzen 9 5950X, and the GPU is a RTX 3090.

The results show that IWPOD-Net trained on NBID from scratch manages a better IoU and competitive OCR score when compared to the other state-of-the-art approaches. IWPOD-Net is shown to be the best model in terms of runtime and performance, which may be thanks to our data augmentation tweaks. 
Not using data augmentation yields a worse IoU score (as seen in Table~\ref{tab:aug_ablation}) than the worst model with respect to the IoU score, YOLO11 in our comparison, which highlights the importance of choosing the best augmentation fit.

Also, as the results show, a higher IoU score does not necessarily improve the fidelity of the OCR for Gemini, but the use of a rectification model always improves the OCR when compared to sending the raw image to the model. 
This means that a rectification method, even if it adds some overhead to the pipeline, is still an important step for a reliable document OCR in this scenario. This also means that the rectification method does not need to be perfect: as long as it leaves all the text visible in the image, a SOTA OCR engine is able to retrieve near-perfect results, meaning a choice for a faster method is often preferable.

\section{Conclusion}
\label{sec:conclusions}

In this work, we presented a novel method for document detection, focusing on the official document scenario, a data augmentation ablation study, cross-dataset evaluation scenarios and a comparison with the SotA.
We also present a novel OCR metric for text retrieval fidelity, and show that our method retains competitive state-of-the-art results while being more time and space efficient.
We contribute by presenting a novel, practical solution to an important real-world problem, and showing how the right experimental tuning can make a smaller network achieve state-of-the-art results.

As future work, the research focus can be expanded to include other document domains such as invoices or receipts, which do not share the assumptions used in this work. This may be done via architectural tweaks over the network, as well as a more thorough cross-dataset examination. Also as future work, we plan on performing a similar study involving real data. There are limitations for using synthetic data for evaluation, especially in terms of OCR fidelity. While this kind of document cannot be made public, it is still possible to reproduce the same experiments and report OCR results for a better overview of the real-world application of these models.

\section*{Acknowledgment}

\iffinal
The authors thank Unico idTech for the support in the making of this research project. This study was financed in part by the \textit{Coordenação de Aperfeiçoamento de Pessoal de Nível Superior - Brasil~(CAPES)}, through the \textit{Programa de Excelência Acadêmica (PROEX)} - Finance Code 001, and in part by the \textit{Conselho Nacional de Desenvolvimento Científico e Tecnológico~(CNPq)} (\# 315409/2023-1). The authors gratefully thank these organizations for the support.
\fi

\bibliographystyle{IEEEtran}

\bibliography{example}

\end{document}